\documentclass[referee,pdflatex,sn-basic,Numbered]{sn-jnl}

\usepackage{graphicx}
\usepackage{amsmath,amssymb,amsfonts}
\usepackage{longtable}
\usepackage{booktabs}

\usepackage{xr}

\makeatletter
\let\origbibcite\bibcite
\let\origcitation\citation
\let\origbibdata\bibdata
\let\origbibstyle\bibstyle

\let\bibcite\@gobbletwo
\let\citation\@gobble
\let\bibdata\@gobble
\let\bibstyle\@gobble

\let\bibcite\origbibcite
\let\citation\origcitation
\let\bibdata\origbibdata
\let\bibstyle\origbibstyle
\makeatother

\begin{document}

\title[MoTIF-X]{MoTIF-X:
A Multimodal Tokenized Framework for
Interpretable and Extensible Molecular Representation Learning}

\author[1]{\fnm{Linqing} \sur{Mo}}

\author*[1,2]{\fnm{Jiayu} \sur{Zhou}}\email{jiayuz@umich.edu}
\author*[1,3,4,5]{\fnm{Bin} \sur{Chen}}\email{chenbi12@msu.edu}

\affil[1]{\orgdiv{Department of Computer Science and Engineering}, \orgname{Michigan State University}, \orgaddress{ \city{East Lansing}, \postcode{48824}, \state{MI}, \country{USA}}}

\affil[2]{\orgdiv{School of Information}, \orgname{University of Michigan}, \orgaddress{\city{Ann Arbor}, \postcode{48109}, \state{MI}, \country{USA}}}

\affil[3]{\orgdiv{Department of Pediatrics and Human Development}, 
          \orgname{Michigan State University}, 
          \orgaddress{\city{Grand Rapids}, \state{MI}, \postcode{49503}, \country{USA}}}

\affil[4]{\orgdiv{Department of Pharmacology and Toxicology}, 
          \orgname{Michigan State University}, 
          \orgaddress{\city{Grand Rapids}, \state{MI}, \postcode{49503}, \country{USA}}}

\affil[5]{\orgdiv{Center for AI-enabled Drug Discovery, College of Human Medicine}, 
          \orgname{Michigan State University}, 
          \orgaddress{\city{Grand Rapids}, \state{MI}, \postcode{49503}, \country{USA}}}

\abstract{
Molecular representation learning is central to computer-aided drug 
discovery, where molecules are described through complementary views 
such as two-dimensional (2D) molecular graphs, motif-level substructures,
simplified molecular-input line-entry system (SMILES) strings, and
three-dimensional (3D) conformations. Although each view captures distinct structural information, integrating them effectively remains challenging. Many existing multimodal approaches learn modality-specific representations independently and align them only at a later stage, restricting fine-grained cross-modal interaction and substructure-level interpretability.

Here, we introduce MoTIF-X, a motif-centered approach that uses molecular motifs as structural anchors for integrating molecular graphs, SMILES strings, and 3D conformational information. Its two-stage pretraining first learns graph-grounded motif representations through hierarchical contrastive learning across atomic, motif, and molecular scales. It then jointly contextualizes motif, SMILES, and discretized torsion-angle tokens through multimodal masked token modeling. Pretrained on a large collection of drug-like molecules with multiple 3D conformers, MoTIF-X achieved the lowest mean absolute error on all nine OpenADMET ExpansionRx endpoints and the best overall performance among the evaluated methods. Significance analyses supported these
improvements in the vast majority of endpoint-baseline comparisons
after multiple-testing correction. Systematic ablation experiments demonstrated the complementary contributions of motif-level token contextualization, multimodal integration, and the two-stage pretraining strategy. When extended to drug-target interaction prediction, MoTIF-X generalized 
to an external drug-cold-start dataset without additional fine-tuning and achieved the best average classification performance across standard and generalization-oriented benchmarks. Furthermore, MoTIF-X supported chemically grounded interpretation at the substructure level: motif attribution scores were positively associated with experimentally measured activity variation, and the model preferentially assigned high attribution to motifs with larger activity shifts.

\par\medskip
\noindent\textbf{Scientific Contribution:}
MoTIF-X advances multimodal molecular representation learning by moving beyond the late-stage alignment of modality-specific embeddings and instead using graph-grounded chemical motifs as cross-modal integration anchors and interpretable molecular units. By combining hierarchical pretraining across chemical scales with the joint contextualization of motif, SMILES, and 3D conformational tokens, MoTIF-X provides a domain-grounded token space for molecular property prediction, drug-target interaction modeling, and substructure-level attribution. Controlled ablations further support the contributions of motif-grounded token organization and cross-modal contextualization
beyond the inclusion of additional downstream modalities alone.
}


\maketitle

\section{Introduction}\label{sec1}

Molecular representations provide a computational basis for computer-aided drug discovery by encoding chemical structures for predictive modeling, molecular analysis and design~\cite{wigh2022review}. Recent advances in artificial intelligence have broadened the use of 
molecular representations beyond task-specific prediction to applications in therapeutic discovery and molecular design~\cite{xing2026deep}. The quality of these representations determines a model's ability to capture chemical properties, generalize across chemical space, and transfer knowledge across tasks. Despite substantial progress, learning representations that are expressive, broadly transferable, and interpretable remains challenging, reflecting the complexity of molecular systems and the need for more unified strategies for organizing molecular information.

This challenge arises in part because molecular structure is described through multiple complementary computational views. Two-dimensional (2D) molecular graphs encode topological connectivity, whereas fragment- or motif-level representations (used interchangeably in this work to denote functional substructures) capture intermediate chemical semantics between atoms and whole molecules. Three-dimensional (3D) conformations describe spatial geometry~\cite{10.24963/ijcai.2023/744}, and simplified molecular-input line-entry system (SMILES) strings provide a symbolic, sequence-based encoding of molecular structure~\cite{mswahili2024transformer}. Other representations, including molecular fingerprints~\cite{yang2022concepts}, experimental measurements and textual annotations, further enrich molecular characterization.

Computational models have been developed to exploit these molecular descriptions individually, including graph-based models for topological connectivity, geometry-aware models for 3D structure and sequence-based models for SMILES encodings~\cite{gilmer2017neural,satorras2021n,schwaller2019molecular}. More recent multimodal approaches have begun to integrate complementary molecular views through graph-geometry alignment, SMILES-graph modeling and the incorporation of motif-level or domain-informed abstractions into graph-based molecular representations~\cite{liu2021pre,wu2023molecular,zhang2021motif,sun2021mocl}.

Despite this progress, several limitations remain. Many existing 
multimodal approaches learn modality-specific representations 
independently and align them only at a later stage. This paradigm embeds different molecular descriptions into separate representation spaces, limiting fine-grained integration of heterogeneous molecular information and joint modeling of chemically coupled features. Moreover, molecular representation frameworks are commonly developed and evaluated for molecular-only tasks and are not explicitly designed to integrate chemical and biological information. Extending a shared molecular representation to cross-domain tasks such as drug-target interaction (DTI) prediction while retaining substructure-level interpretability therefore remains challenging.

To address these limitations, we present MoTIF-X, a motif-centered token integration framework for constructing interpretable and extensible molecular representations. Rather than treating molecular views as independently encoded representations that are aligned only after separate learning, MoTIF-X introduces graph-grounded motif tokens as structural anchors that connect atomic-level chemistry with molecular-level context. The framework first uses hierarchical contrastive learning to ground motif embeddings across atomic, motif and molecular scales. These graph-grounded embeddings are then used to generate motif tokens, which guide the integration of SMILES-derived and 3D conformational tokens within a shared Transformer sequence. This motif-centered design enables integrated molecular representations while preserving substructure-level interpretability.

MoTIF-X achieves the best overall performance among the evaluated
methods on OpenADMET ExpansionRx, with additional molecular property
evaluations on MoleculeNet. It also transfers effectively to DTI prediction under both standard and generalization-oriented settings. Controlled ablations demonstrate the complementary contributions of motif-level token contextualization, multimodal integration, and the two-stage pretraining strategy. The motif-based formulation further supports interpretable analysis, with motif-level attribution capturing experimentally associated activity variation at the substructure level. Together, these findings demonstrate the utility of MoTIF-X as a unified, transferable, and interpretable framework for computational molecular modeling.

\section{Results}\label{sec2}

\subsection{Overview of MoTIF-X}\label{subsec1}

\begin{figure}[!htbp]
    \centering
    \includegraphics[width=1\linewidth]{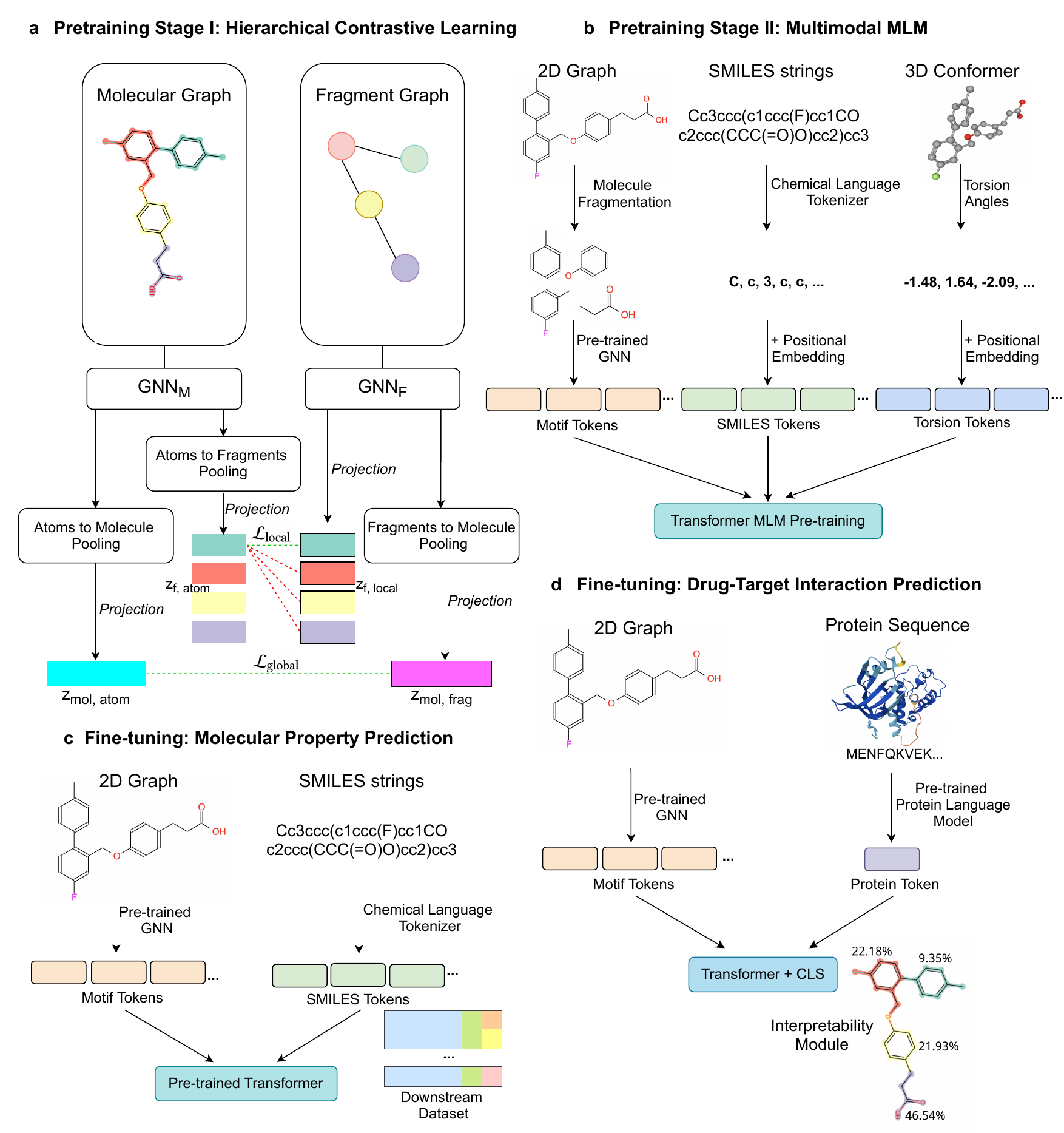}
    \caption{\textbf{Overview of the MoTIF-X framework.}
    \textbf{a} Pretraining Stage I: hierarchical contrastive learning grounds motif representations by aligning atom-, motif-, and molecule-level representations using molecular and fragment graphs, enabling motif representations to bridge atomic detail and molecular context.
    \textbf{b} Pretraining Stage II: multimodal tokenization converts motif, SMILES, and 3D torsion information into tokens, which are jointly modeled using a Transformer with a masked token modeling objective to integrate topological, symbolic, and conformational molecular information. 
    \textbf{c} Downstream molecular property prediction by fine-tuning the pretrained Transformer using motif and SMILES tokens.
    \textbf{d} Downstream drug-target interaction prediction by jointly 
    modeling molecular motif tokens and pretrained protein representations, 
    with attention over motif tokens indicating the substructures emphasized 
    by the model.}
    \label{fig::motifx_overview}
\end{figure}

MoTIF-X is organized around graph-grounded motif tokens that provide structural anchors for integrating SMILES-derived and geometric information. The framework consists of two pretraining stages, followed by downstream applications to molecular property prediction and DTI modeling (Fig.~\ref{fig::motifx_overview}).

The first pretraining stage of MoTIF-X focuses on grounding motif representations across chemical scales. Each molecule is decomposed into a molecular graph and a corresponding fragment graph, where fragments represent chemically meaningful substructures mined from compound libraries (Fig.~\ref{fig::motifx_overview}a). Hierarchical contrastive learning is then applied at both local and global levels. At the local level, atom-level representations pooled within each fragment are aligned with the corresponding fragment representation. At the global level, molecule-level representations derived from molecular and fragment graphs are aligned to promote consistency between local composition and overall molecular context. Together, these objectives enable motif representations to capture intermediate chemical semantics that connect atomic detail with functional substructure context.

The second pretraining stage integrates heterogeneous molecular information into a unified token-based representation for joint modeling (Fig.~\ref{fig::motifx_overview}b). Motif tokens combine motif-pooled atom-level representations from the pretrained molecular graph encoder with corresponding fragment-level representations, yielding tokens grounded in both atomic detail and motif semantics. In addition to motif tokens, SMILES strings and 3D conformational information are represented as sequence-based tokens using a SMILES tokenizer and torsion angle discretization, respectively. All tokens are then jointly processed by a Transformer using a masked token modeling objective~\cite{devlin2019bert}. This stage contextualizes graph-grounded motif tokens together with SMILES and 3D torsion tokens, enabling the Transformer to learn cross-modal dependencies among topological, symbolic and conformational molecular descriptions.

The learned representations are applied to downstream molecular property prediction and DTI modeling (Fig.~\ref{fig::motifx_overview}c,d). For molecular property prediction, motif and SMILES tokens are used during fine-tuning, avoiding explicit 3D conformations that are often computationally expensive to obtain while retaining geometric information captured during pretraining. For DTI prediction, molecular motif tokens are integrated with pretrained protein language representations~\cite{elnaggar2021prottrans} and jointly modeled by the Transformer to capture cross-domain interactions. Beyond predictive performance, the motif-based formulation supports
substructure-level interpretation, as attention over motif tokens
highlights the substructures emphasized by the model during prediction.

\subsection{Molecular Property Prediction Benchmark}\label{property_comparison}

\begin{figure}[!htbp]
    \centering
    \includegraphics[width=\linewidth]{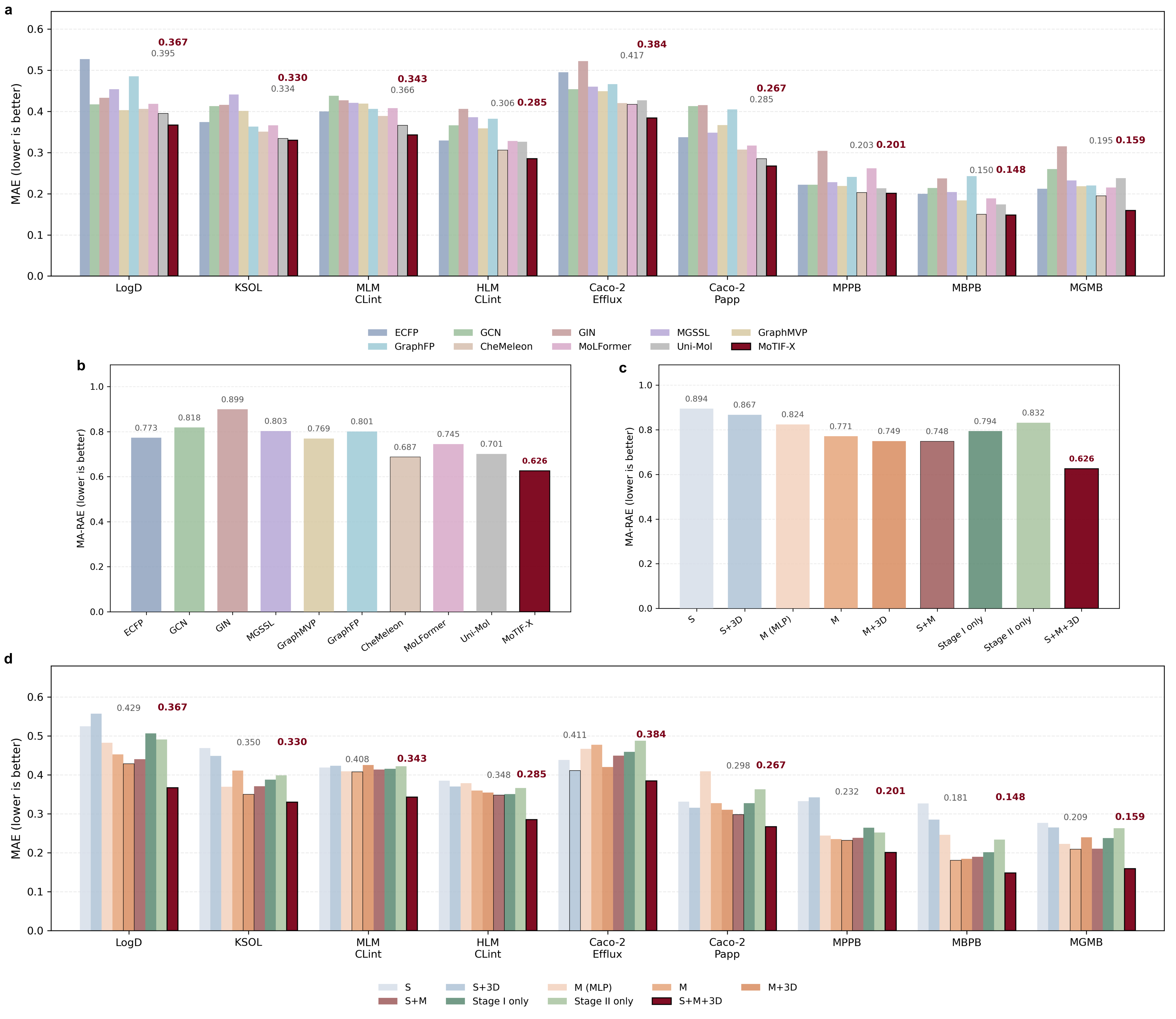}
    \caption{\textbf{MoTIF-X achieves strong molecular property
    prediction performance and benefits from two-stage multimodal
    pretraining on OpenADMET ExpansionRx.}
    \textbf{a-b}, Comparison with nine baseline models.
    MoTIF-X achieves the lowest mean MAE on all nine endpoints
    and the lowest MA-RAE among the compared methods.
    \textbf{a}, Mean absolute error (MAE) across nine ADME endpoints.
    \textbf{b}, Overall performance measured by macro-averaged
    relative absolute error (MA-RAE).
    \textbf{c-d}, Comparison of modality combinations and
    stage-wise pretraining ablations.
    \textbf{c}, MA-RAE across pretraining configurations.
    \textbf{d}, Endpoint-specific MAE across the same configurations.}
    \label{fig:fig2}
\end{figure}

Molecular property prediction serves as a standard testbed for
evaluating the quality and generality of molecular representations.
Within this setting, we evaluate MoTIF-X on OpenADMET
ExpansionRx~\cite{openadmet_expansionrx2026}, covering nine ADME
regression endpoints, and compare against representative approaches
spanning fingerprint-based methods~\cite{rogers2010extended},
graph neural networks (GNNs)~\cite{gilmer2017neural,xu2018powerful},
pretrained molecular models~\cite{zhang2021motif,luong2023fragment,liu2021pre},
SMILES-based Transformers~\cite{ross2022large}, and 3D-aware
molecular models~\cite{zhou2023uni}.
We also evaluate MoTIF-X on MoleculeNet classification and
regression benchmarks~\cite{wu2018moleculenet} to provide
supplementary comparisons across a broader range of molecular
properties (Supplementary Section A.1).
For each benchmark, all methods use identical training, validation,
and test partitions, following the evaluation protocol described
in Section~\ref{evaluation_protocol}.
The relationship between each baseline and MoTIF-X, along with
implementation details, is summarized in Supplementary Section C.1
of Additional file~1.

Across the nine endpoints, MoTIF-X achieves the lowest mean absolute
error (MAE) among all evaluated methods
(Fig.~\ref{fig:fig2}a; Supplementary Table 1).
It also achieves the best overall performance, with a macro-averaged
relative absolute error (MA-RAE) of 0.626, compared with 0.687
for CheMeleon~\cite{burns2025descriptor}, the strongest baseline by this metric
(Fig.~\ref{fig:fig2}b).

The magnitude of improvement over the strongest baseline varies
across tasks.
Larger relative reductions are observed on endpoints such as
LogD and MLM CLint, with smaller numerical differences on
KSOL, MPPB, and MBPB.
Significance analyses further support these improvements, showing
significantly lower MAE for MoTIF-X in the vast majority
of endpoint-baseline comparisons after Holm correction
(Supplementary Table 2).
Together, these results demonstrate a broad predictive advantage
across the benchmark, with gains that vary in magnitude and
statistical significance across endpoints and comparators.

\subsection{Benefits of Two-Stage Multimodal Pretraining for Molecular Property Prediction}\label{ablation_comparison}

Given this strong performance, we next examine which aspects of the MoTIF-X pretraining design contribute to these gains. We compare pretraining configurations that disentangle the effects
of individual molecular views, their combinations, and the two
pretraining stages on OpenADMET ExpansionRx
(Fig.~\ref{fig:fig2}c,d; Supplementary Section A.2).
Overall performance is assessed using MA-RAE, with endpoint-specific
MAEs providing a more detailed comparison. For configurations involving Stage II pretraining, downstream inputs match those used during pretraining, except that 3D conformational tokens are excluded during fine-tuning and inference because reliable conformations are often unavailable in downstream datasets.

In the following comparisons, \emph{S}, \emph{M}, and \emph{3D} denote SMILES, motif, and torsion-angle modalities, respectively, and their combinations indicate the modalities used during pretraining. \emph{S+M+3D} represents the complete MoTIF-X configuration. To assess the benefit of retaining multiple motif tokens,
\emph{M (MLP)} pools all motif representations into a single molecular
representation processed by a multilayer perceptron (MLP).
In contrast, \emph{M} retains individual motif tokens for
Transformer contextualization.

We first assess whether motif-centered tokenization provides a stronger basis for downstream molecular modeling than SMILES-based pretraining. Pretraining with SMILES tokens alone (\emph{S}) yields limited
performance, and adding 3D conformational information in this
SMILES-only setting (\emph{S+3D}) leads to only modest overall
improvement.
By contrast, both motif-based settings achieve better overall
performance, supporting the value of motif-level representations
for downstream learning relative to SMILES-based alternatives. Beyond motif extraction, retaining motifs as individual tokens
provides a further benefit: \emph{M} outperforms \emph{M (MLP)}
overall and on most endpoints, supporting the use of token-level
contextualization.

Building on the motif-centered setting, we next ask whether
incorporating complementary molecular views yields further gains.
Relative to \emph{M}, all multimodal variants, including
\emph{M+3D}, \emph{S+M}, and \emph{S+M+3D}, achieve stronger
overall performance.
To assess whether these gains can be explained by the inclusion
of additional downstream inputs alone, we introduce
\emph{Stage I only} as a control.
This configuration uses the same motif and SMILES inputs as the
complete MoTIF-X model and initializes the graph encoders through
Stage I, but omits Stage II pretraining; the token embeddings,
Transformer, and prediction head are therefore randomly initialized
before downstream fine-tuning.
\emph{Stage I only} performs at a similar level to \emph{M},
suggesting that the addition of SMILES inputs alone does not
account for the observed gains.
In contrast, the multimodal variants pretrained during Stage II
achieve better overall performance than both configurations.
Together, these comparisons support the contribution of joint
multimodal pretraining beyond simply increasing the number of
downstream input modalities.

We next assess whether Stage II alone can recover these benefits
without the graph-grounded initialization provided by Stage I.
In the \emph{Stage II only} configuration, multimodal Stage II
pretraining is performed without Stage I: motif tokens are generated
by randomly initialized, frozen graph encoders, while the token
embeddings and Transformer are trained.
\emph{Stage II only} achieves better overall performance than
\emph{S+3D} but remains below \emph{M}, with the same ordering
observed on most endpoints.
This pattern suggests that motif decomposition contributes useful
structural information even without Stage I initialization,
but that graph-grounded initialization helps the model make
more effective use of motif-token representations.

Overall, the strongest performance arises from the full MoTIF-X design, which combines motif-centered token modeling with multimodal information integration: Stage I establishes graph-grounded motif representations, while Stage II further contextualizes them with SMILES and 3D information to improve downstream property prediction.

\subsection{Multimodal Pretraining Reshapes Attention and Representation Geometry}

\begin{figure}[!htbp]
    \centering
    \includegraphics[width=1\linewidth]{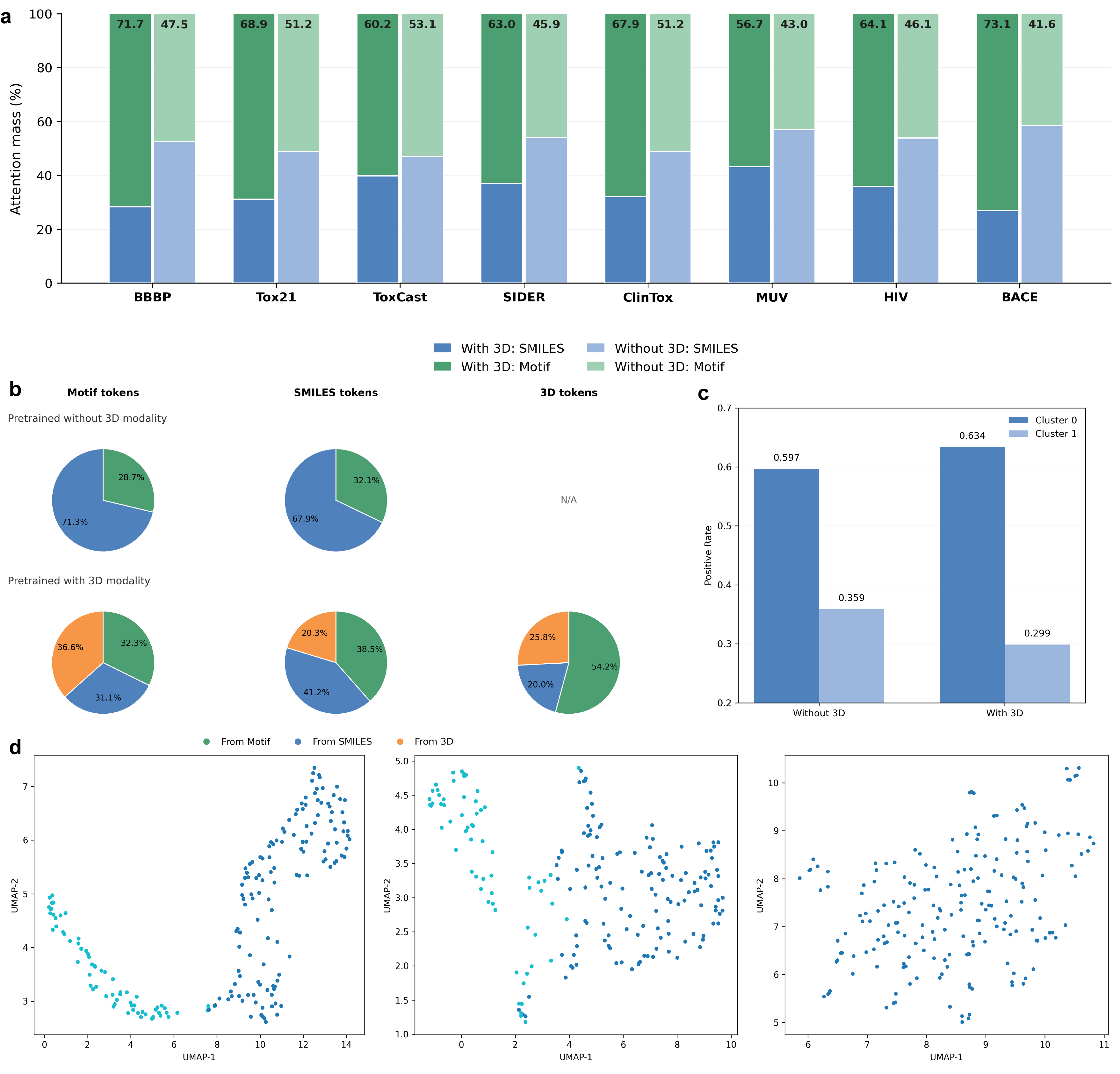}
    \caption{\textbf{Multimodal pretraining redistributes attention across token types and shapes representation geometry in MoTIF-X.}
    \textbf{a-b}, Attention redistribution across token types. Pretraining with 3D tokens shifts downstream attention toward motif tokens and promotes interactions among motif, SMILES, and 3D tokens during pretraining.
    \textbf{a}, Incoming attention received by the classification token 
    (\texttt{[CLS]}) from motif and SMILES tokens during downstream prediction across MoleculeNet benchmarks.
    \textbf{b}, Final-layer incoming attention composition for motif, SMILES, and 3D target tokens during pretraining.
    Top: models pretrained without 3D information. Bottom: models pretrained with 3D information.
    \textbf{c-d}, Representation geometry induced by multimodal pretraining. Multimodal pretraining yields more structured and separable representations.
    \textbf{c}, Cluster-level positive rate on BBBP, showing stronger alignment between embedding clusters and downstream labels after pretraining with 3D tokens.
    \textbf{d}, Uniform manifold approximation and projection (UMAP) of 
    learned \texttt{[CLS]} representations on BBBP. From left to right: 
    motif+SMILES+3D, motif+SMILES, and extended-connectivity fingerprint 
    (ECFP) representations.}
    \label{fig:fig3}
\end{figure}

To characterize how multimodal pretraining alters molecular
representations, we analyze token-level attention during pretraining
and examine downstream attention patterns and learned
\texttt{[CLS]} embeddings on MoleculeNet benchmarks
(Fig.~\ref{fig:fig3}; Supplementary Section D). During pretraining, we quantify the incoming attention received by each token type from different source token types. Without 3D information (MoTIF-X (S+M)), attention is largely concentrated on SMILES tokens, indicating greater
attention allocation to SMILES-derived sequence information. Incorporating 3D tokens (MoTIF-X (S+M+3D)) redistributes attention toward motif and 3D sources, reducing the dominance of SMILES self-attention and promoting more balanced interactions among motif, SMILES and 3D tokens (Fig.~\ref{fig:fig3}b).

These effects persist during downstream prediction. The attention received by the \texttt{[CLS]} token, which provides an attention-based view of token contributions to the final prediction, shifts toward motif tokens and away from SMILES tokens across tasks when the model is pretrained with 3D information (Fig.~\ref{fig:fig3}a). This pattern is consistent with geometric information acquired during
pretraining influencing the downstream utilization of motif-level
representations.

We next ask whether this shift in token utilization is reflected in embedding geometry. Using BBBP as an example, cluster-level analysis of CLS embeddings shows that incorporating 3D information yields more compact and better separated clusters than pretraining without 3D. This more organized structure is also more closely aligned with the downstream task: clusters from the 3D-pretrained model show a larger difference in positive-class prevalence between clusters than those from the model without 3D pretraining (Fig.~\ref{fig:fig3}c). These cluster-level trends are consistent across most MoleculeNet classification benchmarks (Supplementary Tables 18, 19). The main exception is Toxicology in the 21st Century (Tox21), where weaker cluster-level label separation coincides with relatively lower predictive performance (Supplementary Table 3). UMAP visualization provides a qualitative view of the same pattern, with 3D-pretrained representations forming a more structured manifold and ECFP features showing a more diffuse distribution (Fig.~\ref{fig:fig3}d).

Together, these analyses indicate that multimodal pretraining reshapes both attention allocation and representation geometry, shifting MoTIF-X toward motif-centered, geometry-enriched representations that better align with downstream prediction signals.

\subsection{MoTIF-X Extends to Drug-Target Interaction Prediction}\label{dti_intro}

\begin{figure}[!htbp]
    \centering
    \includegraphics[width=1\linewidth]{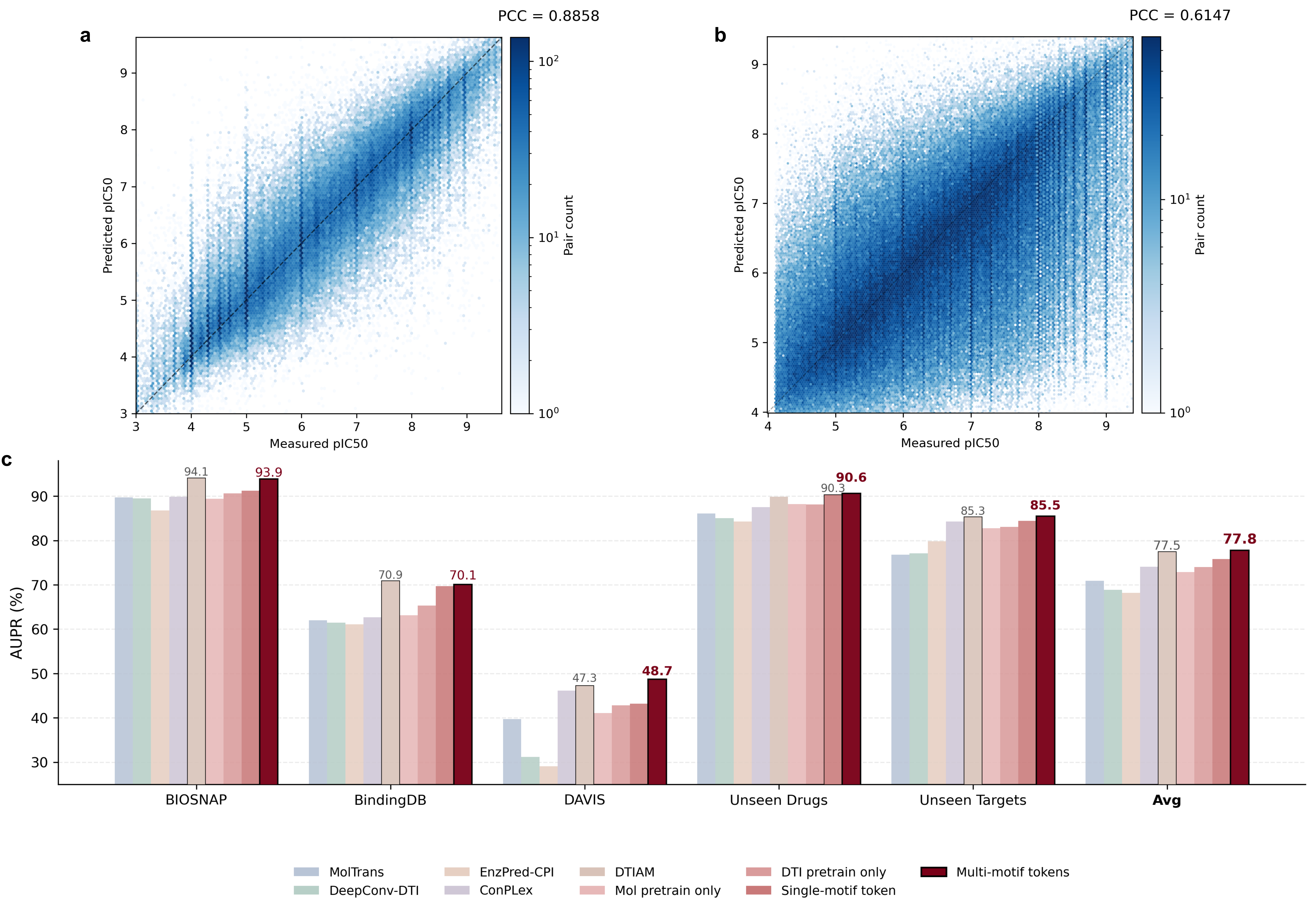}
    \caption{\textbf{MoTIF-X transfers to drug-target interaction prediction.}
    \textbf{a}, In-domain half-maximal inhibitory concentration (IC$_{50}$) 
    regression on the BindingDB test set, shown as predicted versus measured 
    pIC$_{50}$ values, where pIC$_{50}$ denotes the negative base-10 logarithm 
    of the molar IC$_{50}$ value. Pearson correlation coefficient (PCC) 
    indicates agreement with measured binding affinities.
    \textbf{b}, External evaluation under a drug cold-start setting on a ChEMBL IC$_{50}$ dataset without additional fine-tuning, evaluated by PCC, indicating generalization to previously unseen compounds across datasets.
    \textbf{c}, DTI classification performance across multiple benchmarks, 
    reported as area under the precision-recall curve (AUPR; \%) with 
    comparison to baseline methods.
    \emph{Mol pretrain only} uses pretrained molecular embeddings, while the Transformer is initialized randomly before downstream fine-tuning and no DTI-specific pretraining is applied.
    \emph{DTI pretrain only} initializes the molecular encoders randomly before DTI-specific pretraining, rather than using encoders initialized from molecular pretraining.
    \emph{Single-motif token} and \emph{Multi-motif tokens} both use pretrained molecular embeddings and DTI-specific pretraining, with \emph{Multi-motif tokens} denoting the standard MoTIF-X formulation that represents each molecule using multiple motif tokens.
    MoTIF-X achieves the highest average AUPR and performs best on generalization-oriented settings.}
    \label{fig:fig4}
\end{figure}

Molecular property prediction benchmarks provide a useful yet incomplete evaluation of molecular representations, as they focus primarily on intrinsic properties of individual molecules. To assess interaction-level modeling, we extend MoTIF-X to DTI prediction, where the model must reason over interactions between small-molecule compounds and protein targets. Unlike task-specific DTI architectures that couple separately designed drug and protein encoders through late-stage fusion~\cite{ozturk2018deepdta, lee2019deepconv, chen2020transformercpi, huang2021moltrans, nguyen2021graphdta}, MoTIF-X represents molecules as chemically meaningful motif-token sequences shared across tasks. In the DTI setting, these molecular motif tokens are jointly modeled with ProtBERT-derived protein representations~\cite{elnaggar2021prottrans} in a unified Transformer, enabling interaction prediction while preserving the generality and interpretability of the learned molecular representation.

We first evaluate MoTIF-X on large-scale DTI activity regression using a BindingDB-based IC$_{50}$ dataset (Sections~\ref{method_datasets}, \ref{evaluation_protocol}). On the BindingDB test set, MoTIF-X achieves a Pearson correlation coefficient (PCC) of 0.8858, showing strong agreement with measured binding affinities (Fig.~\ref{fig:fig4}a). To assess drug cold-start transfer in an external dataset, we directly evaluate the BindingDB-trained model on a ChEMBL IC$_{50}$ dataset after excluding any ChEMBL compounds appearing in the BindingDB training set. MoTIF-X achieves a PCC of 0.6147 without additional fine-tuning (Fig.~\ref{fig:fig4}b), demonstrating measurable cross-dataset transfer to compounds unseen during training. Given these in-domain and external drug cold-start results, this BindingDB-pretrained checkpoint is used as the common initialization for subsequent DTI fine-tuning and model comparisons (Section~\ref{downstream_finetuning_overview}).

We further evaluate MoTIF-X on widely used DTI classification benchmarks, including the Stanford Biomedical Network Dataset Collection 
(BIOSNAP)~\cite{biosnapnets}, BindingDB~\cite{liu2007bindingdb}, and DAVIS~\cite{davis2011comprehensive}, together with two BIOSNAP-derived generalization settings, \emph{Unseen Drugs} and \emph{Unseen Targets} (Section~\ref{method_datasets}; Fig.~\ref{fig:fig4}c; Supplementary Table 9). Across these datasets and settings, MoTIF-X is compared with representative DTI baselines spanning convolutional, transformer-based, protein-informed, contrastive, and attention-based architectures~\cite{lee2019deepconv,huang2021moltrans,goldman2022machine,singh2023contrastive,lu2025dtiam}. The complete MoTIF-X model achieves the highest average AUPR, with the strongest gains on DAVIS, \emph{Unseen Drugs}, and \emph{Unseen Targets}. Statistical comparisons with the baseline methods are provided in Supplementary Table 10. These settings place greater emphasis on generalization, either by evaluating unseen drugs or targets, or, in DAVIS, by requiring discrimination among kinase interactions with similar binding pockets and limited training data~\cite{davis2011comprehensive}. This motivates examining how molecular pretraining, DTI-specific pretraining, and motif-level granularity contribute to these gains.

Controlled DTI variants further clarify the sources of improvement (Fig.~\ref{fig:fig4}c). Both \emph{Mol pretrain only} and \emph{DTI pretrain only} fall short of the complete MoTIF-X model, showing that the full gains do not arise from molecular pretraining or DTI-specific pretraining alone, but from their combination. Between these two reduced variants, \emph{DTI pretrain only} generally remains closer to the complete model across benchmarks, indicating the importance of task-aligned interaction pretraining. By contrast, the clearest relative benefit of molecular pretraining appears in the more generalization-oriented \emph{Unseen Drugs} setting, consistent with its role in drug-side generalization. Finally, \emph{Multi-motif tokens} outperforms \emph{Single-motif token}, indicating that preserving motif-level granularity improves drug-protein interaction modeling. Together, these results indicate that MoTIF-X naturally extends to cross-domain DTI prediction, with motif-level tokens supporting both generalization and interaction-aware modeling.

\subsection{Motif-level Attribution Reflects Experimental Activity Variation}\label{dti_experimental_variation}

The strong predictive performance of MoTIF-X across the evaluated DTI benchmarks demonstrates its utility for DTI modeling, while its motif-token architecture also supports attribution at the level of chemically meaningful substructures. As individual molecular tokens correspond to motifs, final-layer attention from the learnable \texttt{[CLS]} token to motif tokens provides an attention-based substructure-level attribution signal. We use this signal to examine whether model attributions are consistent with experimentally observed activity variation. Specifically, we perform a motif-centered analysis of experimentally measured pIC$_{50}$ values from ChEMBL for signal transducer and activator of transcription 3 (STAT3)-targeting compounds (Supplementary Section E.1) using a STAT3-specific model trained as described in Supplementary Section E.3. For each motif, we calculate a motif-associated activity shift ($\Delta pIC_{50}$), defined as the difference in mean pIC$_{50}$ between molecules with and without the motif, and compare its magnitude with mean motif attribution. Across all motifs occurring in at least 10 molecules, motif attribution is moderately correlated with the magnitude of the observed activity shift (Spearman $\rho = 0.44$ for $|\Delta pIC_{50}|$), indicating that motifs associated with larger activity differences tend to receive higher attribution scores.

We next examine whether motifs associated with larger experimental activity differences are preferentially prioritized within individual molecules. Motifs are first ranked globally by $|\Delta pIC_{50}|$, and different fractions of the top-ranked motifs are selected. For each ratio, we compute a Top-$k$ hit rate, defined as the proportion of selected motif-molecule pairs for which the motif is ranked among the top-$k$ motifs within the corresponding molecule according to model attribution (Fig.~\ref{fig:fig5}a). Stronger enrichment is observed at lower selection ratios, indicating that motifs with larger experimentally observed activity shifts are more likely to receive high attribution scores. At a 0.05 selection ratio, the Top-1 hit rate is enriched by approximately 2.55$\times$ relative to random selection, and the Top-2 hit rate shows a consistent enrichment trend. Consistent with this enrichment analysis, the top 20 motifs ranked by $|\Delta pIC_{50}|$ also receive high attribution scores and are frequently prioritized within individual molecules (Fig.~\ref{fig:fig5}b; Supplementary Section E.4). These motifs include both activity-enhancing and activity-reducing substructures, indicating that MoTIF-X highlights motifs associated with both increased and decreased binding activity.

Together, these results show that motif-level attribution in MoTIF-X is consistent with experimentally observed activity variation. To further assess the consistency of attention-based attribution, we compare it with Input$\times$Gradient~\cite{shrikumar2017learning}, a commonly used gradient-based attribution method. The two approaches show strong agreement across motif instances in the STAT3 dataset (Pearson $r = 0.881$; Supplementary Section E.2), demonstrating methodological agreement between the two attribution approaches. These findings motivate further examination of whether attribution signals can yield chemically informative insight into drug-target interactions in a disease-relevant setting.

\subsection{Motif-level Attribution Highlights Chemically Informative Substructures in Drug-Target Interaction Prediction}\label{dti_mechanism}

\begin{figure}[!htbp]
    \centering
    \includegraphics[width=1\linewidth]{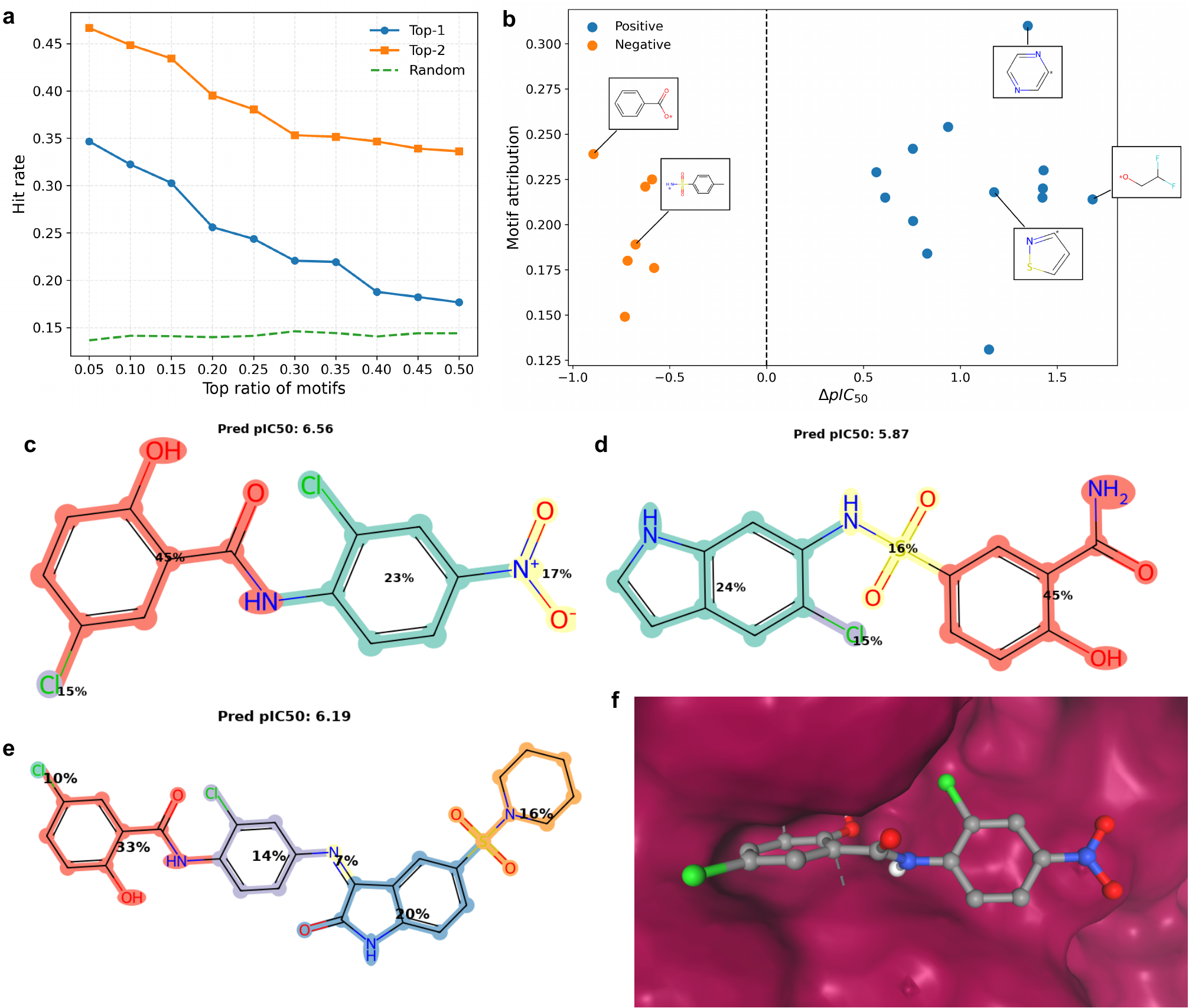}
    \caption{\textbf{Motif-level attribution in MoTIF-X aligns with activity variation and highlights chemically informative substructures.}
    \textbf{a-b}, Motif-level attribution aligns with experimental activity variation, with activity-relevant motifs receiving higher attribution and being preferentially ranked by the model. 
    \textbf{a}, Top-$k$ hit rate versus motif selection ratio. Higher values at lower ratios indicate that the most activity-relevant motifs are preferentially ranked.
    \textbf{b}, Examples of top-ranked motifs plotted by activity change ($\Delta pIC_{50}$) and motif attribution. MoTIF-X assigns high attribution to motifs associated with both increased and decreased activity and frequently prioritizes them within molecules.
    \textbf{c-e}, Motif-level attribution identifies localized and consistent substructure-level signals across structurally diverse hepatocellular carcinoma (HCC)-related compounds, with the Salicylamide motif highlighted in orange-red and consistently receiving high attribution.
    \textbf{c}, Niclosamide against STAT3.
    \textbf{d-e}, Additional HCC-related compounds showing a similar attribution pattern despite structural variation.
    \textbf{f}, Structural plausibility of motif-level attribution examined by molecular docking. The top-ranked predicted Niclosamide-STAT3 pose yields a favorable docking score, with predicted protein-ligand contacts involving the Salicylamide motif.}
    \label{fig:fig5}
\end{figure}

We next examine whether motif-level attribution can highlight chemically informative substructures in a disease-relevant DTI setting. This setting is particularly relevant to practical drug discovery, where predictive accuracy alone is often insufficient for rational lead optimization and attribution can help identify chemical motifs associated with predicted interactions~\cite{arrowsmith2015promise}. We therefore use a case study to assess whether MoTIF-X identifies chemically plausible motif-level attribution patterns and whether these patterns recur across related compounds and a larger chemical space.

The case study centers on Niclosamide, a clinically approved anthelmintic previously implicated in transcriptomic and systems-level studies of hepatocellular carcinoma (HCC)~\cite{li2014multi, chen2017computational, chen2017reversal}. To place the analysis in a biologically informed context, we evaluate it against a curated HCC-associated pathway protein set rather than the full proteome (Supplementary Section F.3). Within this protein space, MoTIF-X produces a selective target profile for Niclosamide, with STAT3 and inhibitor of nuclear factor kappa-B kinase subunit beta 
(IKBKB) emerging as the top-ranked predicted targets and showing higher predicted interaction strengths than other pathway proteins. For the Niclosamide-STAT3 pair, MoTIF-X predicts a strong interaction (pIC$_{50}$ = 6.56; IC$_{50} \approx 0.28~\mu$M; Fig.~\ref{fig:fig5}c). The Salicylamide motif receives the highest motif-level attribution, identifying it as the substructure most strongly emphasized by the model for this prediction. Additional HCC-related compounds~\cite{omran2024pharmacological, WO2017153952A1} show a similar model-attribution pattern: despite broader structural variation, MoTIF-X again assigns high attribution to the Salicylamide motif in compounds predicted to interact with STAT3 (Fig.~\ref{fig:fig5}d,e).

To test whether this pattern extends to a broader chemical space, we computationally screen a 430k-molecule ChEMBL diversity subset~\cite{mayr2018large} against STAT3. The predicted interaction scores show a broad but selective distribution, with high-scoring compounds concentrated in a small fraction of the library (Supplementary Section F.1). Motif enrichment analysis of the top-ranked molecules identifies 798 unique motifs across the dataset, among which the Salicylamide motif ranks second by enrichment score (Supplementary Section F.2). This indicates that the motif highlighted in the Niclosamide case is not isolated but forms part of a broader structural pattern enriched among top-ranked compounds in the screening, consistent with MoTIF-X capturing recurrent substructure-level signals associated with predicted STAT3 interactions.

We finally examine whether the highlighted motif is structurally plausible in a predicted binding pose. Molecular docking with AutoDock Vina~\cite{eberhardt2021autodock} yields a top-ranked Niclosamide-STAT3 pose with a docking score of $-6.702$ kcal/mol. In this pose, the predicted protein-ligand contacts involve atoms within the Salicylamide motif (Fig.~\ref{fig:fig5}f; Supplementary Sections F.4, F.5). This result provides a complementary structure-based consistency check for the motif-level attribution produced by MoTIF-X.

Taken together, these results show that MoTIF-X identifies motif-level attribution patterns that are consistent across individual compounds, structurally related candidates, and large-scale chemical libraries. By highlighting substructures that are repeatedly assigned high attribution in predicted drug-target interactions, the model provides a structure-grounded view of the chemical motifs it emphasizes during prediction. This motif-centered view may therefore help prioritize substructures for subsequent experimental evaluation during lead optimization and analog design.

\section{Discussion}\label{discussion}

This study proposes MoTIF-X, a motif-centered shared-token framework for multimodal molecular representation learning. MoTIF-X uses graph-grounded motif tokens as structural anchors for jointly contextualizing motif, SMILES, and 3D torsion information within a shared Transformer, and extends this formulation to DTI prediction by integrating molecular motif tokens with protein representations. The framework was evaluated on OpenADMET ExpansionRx and DTI
benchmarks, with supplementary molecular property comparisons
on MoleculeNet. Additional analyses included controlled ablations, attention and representation geometry, motif-level attribution,
and an HCC-focused computational case study. Together, these results show that structurally grounded tokenization can improve predictive performance, support transfer across task settings, and enable chemically interpretable analysis.

Across these evaluations, performance gains arise not simply from adding more modalities, but from organizing molecular information at an appropriate structural granularity. Motif-level tokens provide an intermediate scale between atomic detail and whole-molecule embeddings, preserving compositional structure while enabling explicit substructure interactions. The ablations further show that both pretraining stages contribute to this process: hierarchical contrastive pretraining grounds motifs in atomic and molecular context, whereas multimodal masked token modeling contextualizes these motifs with SMILES and 3D torsion tokens. This combination supports robust molecular property prediction and extends naturally to DTI modeling, where protein binding often depends on specific combinations of functional groups rather than a single global molecular summary.

The attention and representation analyses further clarify how this organization changes multimodal learning. Rather than combining independently formed modality-specific representations only at a final stage, MoTIF-X allows heterogeneous molecular views to interact within a shared token sequence. The observed redistribution of attention across token types and the accompanying changes in \texttt{[CLS]} embedding geometry indicate that multimodal pretraining alters both attention allocation and representation organization. Together with the ablation results, these observations support the view that the benefits of MoTIF-X arise from learned contextualization among motif, SMILES, and 3D information rather than from simply adding or aligning separate molecular views.

Despite these strengths, several limitations merit consideration. First, although geometric information is incorporated during pretraining, explicit 3D representations are not required at inference time. While this improves practical applicability, tasks that depend strongly on spatial configuration, such as energetic and thermodynamic property prediction~\cite{ramakrishnan2014quantum}, remain to be systematically evaluated in settings that incorporate explicit 3D information during downstream inference. Second, although motif-level attribution and attention analyses provide interpretable signals, these explanations remain model-derived and do not constitute direct mechanistic evidence of molecular binding. The STAT3 attribution analysis, virtual screening, and molecular docking therefore provide computational consistency evidence rather than experimental confirmation of binding affinity or mechanism. Prospective biochemical and structure-activity experiments will be required to validate the highlighted targets and substructures.

Looking forward, the token-based formulation of MoTIF-X offers a flexible foundation for extension. Additional modalities, such as textual annotations, physicochemical descriptors, or other structured signals, could be incorporated within the same token space without fundamental redesign. The motif-centered representation may also support generative and molecular modification tasks, including motif-level editing, controllable generation, and structure-guided optimization.

\section{Conclusions}\label{conclusions}

In this work, we introduced MoTIF-X, a motif-centered framework that uses graph-grounded chemical motifs to integrate molecular graphs, SMILES strings, and 3D conformational information within a shared Transformer representation. MoTIF-X achieved consistently strong performance across molecular property prediction and drug-target interaction tasks, including external and generalization-oriented evaluations. Controlled ablations 
indicated that the performance gains arose from motif-centered organization and cross-modal contextualization rather than from the addition of modalities alone. Motif-level attribution was also associated with 
experimentally observed activity variation, linking model predictions to chemically meaningful substructures. Together, these findings demonstrate how heterogeneous descriptions of a molecular system can be organized 
within a shared, domain-grounded token space while preserving transferability and chemical interpretability.

\section{Methods}\label{sec4}

\subsection{Model architecture}\label{model_architecture_overview}

MoTIF-X consists of two sequential pretraining stages followed by 
task-specific fine-tuning. The molecular and fragment encoders are 
implemented using Graph Isomorphism Networks (GINs)~\cite{xu2018powerful}, 
with five and two message-passing layers, respectively, and a hidden 
dimension of 300. Molecular graphs use categorical physicochemical and 
stereochemical atom and bond features, whereas fragment graphs use 
motif-vocabulary identifiers as node features and untyped bidirectional 
adjacency edges. A six-layer Transformer with six attention heads and 
the same hidden dimension processes the heterogeneous token sequence. 
Stage I aligns atomic, motif, and molecular representations through 
contrastive learning, whereas Stage II performs multimodal masked token 
modeling.

\subsubsection{Stage I: Hierarchical Contrastive Pretraining}

The local objective grounds motif representations in their constituent atoms, whereas the global objective aligns fragment-based and atom-level molecular representations. 

\paragraph{Molecule Fragmentation}

To construct the motif graph used in Stage I pretraining, we apply the 
Principal Subgraph Mining algorithm~\cite{kong2022molecule,luong2023fragment} 
to derive a motif vocabulary from the Stage I pretraining molecules. The 
resulting vocabulary is fixed for all subsequent pretraining and downstream 
datasets.

Using this vocabulary, each molecule is partitioned into disjoint motifs 
that collectively cover the molecular graph. A fragment graph is then 
constructed in which each node represents a motif, and two nodes are 
connected when their corresponding motifs are adjacent in the original 
molecular graph. The resulting fragment graph is used as input to the 
fragment-level encoder.

\paragraph{Local-level Contrastive Learning}

A local-level contrastive objective is employed to align atom-level and fragment-level representations, encouraging each motif embedding to preserve the chemical information of its constituent atoms. 
The atom-level representation is computed by $\mathrm{GNN}_\mathrm{M}$ on the molecular graph, while the fragment-level representation is computed by $\mathrm{GNN}_\mathrm{F}$ on the fragment graph.

Given a molecular graph $G_M=(V_M,E_M)$ and its fragment graph $G_F=(V_F,E_F)$, 
node-level atom embeddings and fragment-node embeddings are computed as
\begin{align}
    \mathbf{h}_{\text{atom}} &= \mathrm{GNN}_\mathrm{M}(V_M,E_M), \\
    \mathbf{h}_{\text{frag}} &= \mathrm{GNN}_\mathrm{F}(V_F,E_F).
\end{align}

Atom embeddings are aggregated within each fragment using mean pooling based on the atom-to-fragment mapping:
\begin{align}
    \mathbf{h}_{\text{f,atom}} = \mathrm{MEANPOOL}(\mathbf{h}_{\text{atom}}, \text{atom-to-fragment mapping}).
\end{align}

Both $\mathbf{h}_{\text{f,atom}}$ and $\mathbf{h}_{\text{frag}}$ are projected into a shared latent space via a local projection head $f_{\text{local}}(\cdot)$ (a two-layer MLP) followed by $\ell_2$ normalization:
\begin{align}
    \mathbf{z}_{\text{f,atom}} &= \frac{f_{\text{local}}(\mathbf{h}_{\text{f,atom}})}{\| f_{\text{local}}(\mathbf{h}_{\text{f,atom}})\|_2}, \\
    \mathbf{z}_{\text{f,local}} &= \frac{f_{\text{local}}(\mathbf{h}_{\text{frag}})}{\| f_{\text{local}}(\mathbf{h}_{\text{frag}})\|_2}.
\end{align}

Negative samples are drawn from other fragments within the same 
mini-batch. The local contrastive loss is defined as
\begin{align}
\mathcal{L}_{\text{local}}
=
-\frac{1}{B_f}\sum_{i=1}^{B_f}
\log
\frac{
\exp\left(
\langle
\mathbf{z}^{(i)}_{\text{f,atom}},
\mathbf{z}^{(i)}_{\text{f,local}}
\rangle/\tau
\right)
}{
\sum_{j=1}^{B_f}
\exp\left(
\langle
\mathbf{z}^{(i)}_{\text{f,atom}},
\mathbf{z}^{(j)}_{\text{f,local}}
\rangle/\tau
\right)
},
\end{align}
where $B_f$ is the number of fragments in the mini-batch.

\paragraph{Global-level Contrastive Learning}

A global-level contrastive objective is defined to align molecule-level representations derived from atom-level and fragment-level embeddings, encouraging the fragment-based representation to capture how local motifs are organized within the complete molecular graph.

Molecule-level representations are obtained via mean pooling:
\begin{align}
    \mathbf{h}_{\text{mol,atom}} &= \mathrm{MEANPOOL}(\mathbf{h}_{\text{atom}}, \text{atom-to-molecule mapping}), \\
    \mathbf{h}_{\text{mol,frag}} &= \mathrm{MEANPOOL}(\mathbf{h}_{\text{frag}}, \text{fragment-to-molecule mapping}).
\end{align}

These representations are projected into a shared latent space using a global projection head $f_{\text{global}}(\cdot)$ (a two-layer MLP) followed by $\ell_2$ normalization:
\begin{align}
    \mathbf{z}_{\text{mol,atom}} &= \frac{f_{\text{global}}(\mathbf{h}_{\text{mol,atom}})}{\| f_{\text{global}}(\mathbf{h}_{\text{mol,atom}})\|_2}, \\
    \mathbf{z}_{\text{mol,frag}} &= \frac{f_{\text{global}}(\mathbf{h}_{\text{mol,frag}})}{\| f_{\text{global}}(\mathbf{h}_{\text{mol,frag}})\|_2}.
\end{align}

Negative samples are drawn from other molecules within the same 
mini-batch. The global contrastive loss is defined as
\begin{align}
\mathcal{L}_{\text{global}}
=
-\frac{1}{B_m}\sum_{i=1}^{B_m}
\log
\frac{
\exp\left(
\langle
\mathbf{z}^{(i)}_{\text{mol,atom}},
\mathbf{z}^{(i)}_{\text{mol,frag}}
\rangle/\tau
\right)
}{
\sum_{j=1}^{B_m}
\exp\left(
\langle
\mathbf{z}^{(i)}_{\text{mol,atom}},
\mathbf{z}^{(j)}_{\text{mol,frag}}
\rangle/\tau
\right)
},
\end{align}
where $B_m$ is the number of molecules in the mini-batch.

The overall Stage I objective is defined as
\begin{align}
    \mathcal{L}_{\text{Stage I}}
    =
    \mathcal{L}_{\text{local}}
    +
    \lambda\mathcal{L}_{\text{global}},
\end{align}
where $\lambda \in \mathbb{R}_{+}$ controls the relative contribution of 
the global objective. In practice, both the temperature $\tau$ and the 
global-loss weight $\lambda$ are set to 0.1 based on preliminary 
hyperparameter tuning.

Together, these objectives provide a graph-grounded initialization for 
motif-centered representations before multimodal masked token modeling 
in Stage II.

\subsubsection{Stage II: Multimodal Masked Token Modeling}

Stage II constructs motif, SMILES, and torsion angle tokens and jointly trains them within a unified Transformer using a masked token modeling objective.

\paragraph{Motif Tokens}

Using the pretrained and frozen graph encoders from Stage I, the token 
representation of motif $k$ is constructed by combining its pooled 
atom-level representation with the corresponding fragment-level 
representation:
\begin{align}
    \mathbf{h}_{\text{motif}}^{(k)}
    =
    \mathbf{h}_{\text{f,atom}}^{(k)}
    +
    \mathbf{h}_{\text{frag}}^{(k)}.
\end{align}
The resulting motif tokens are used as graph-grounded inputs to the 
Transformer in Stage II.

\paragraph{SMILES Tokens}

SMILES tokens are generated from each canonical SMILES string using a 
case-sensitive greedy longest-match tokenizer. At each position, the 
longest matching entry in the shared sequence vocabulary is selected, 
whereas unmatched characters are retained as single-character tokens. 
The resulting tokens are embedded using a learnable token embedding matrix.

\paragraph{Torsion Angle Tokens}

Torsion angle tokens are constructed from the molecular conformations described in Supplementary Section B.1. 
For each conformer, all valid dihedral angles associated with rotatable bonds are enumerated as ordered atom quadruples $(i,a,b,l)$. 
To ensure a consistent torsion sequence across conformers of the same molecule, the torsions are ordered according to a depth-first search traversal of the molecular graph.

The continuous dihedral angles are computed in radians and discretized over the range $[-\pi,\pi]$ using a uniform bin size of $0.01$ radians. Each discretized value is mapped to an integer identifier using entries reserved for torsion-angle bins in the sequence-token vocabulary. Although the same vocabulary also includes SMILES-symbol entries, these entries are disjoint from the torsion-angle-bin entries. Thus, the shared vocabulary provides a common indexing and embedding mechanism while keeping SMILES and torsion tokens as separate token types.

The resulting torsion token sequence provides a topology-consistent encoding of molecular conformation. 
These tokens are embedded using the shared token embedding matrix and combined with fixed sinusoidal positional encodings before being processed jointly with motif and SMILES tokens in the Transformer.

The multimodal input concatenates motif, SMILES, and torsion tokens in 
this order and is right-padded or right-truncated to a maximum length of 
200 before the \texttt{[CLS]} token is prepended.

\paragraph{Masked Token Modeling Objective}

Let $\mathbf{x}=(x_1,\ldots,x_N)$ denote the concatenated motif, SMILES, 
and torsion token sequence. During training, 15\% of non-padding positions 
are selected as prediction targets, with at least one position selected 
from each available modality. Selected SMILES and torsion tokens follow 
the standard 80/10/10 replacement strategy. At selected motif positions, 
the graph-derived motif embeddings remain as inputs and are used to 
predict the corresponding motif identities.

A token prediction head is applied to the representations at selected positions to predict the original token identities. 
For motif tokens, prediction is optimized using a standard cross-entropy loss over the motif vocabulary generated during molecule fragmentation. 
For SMILES tokens, prediction is optimized using cross-entropy over the SMILES-symbol entries in the sequence-token vocabulary. 
For torsion tokens, prediction over the torsion-angle-bin entries in the same sequence-token vocabulary is optimized using the Gaussian cross-entropy (GCE) loss described in~\cite{wang2025token}.

The Stage II objective is the sum of the motif cross-entropy, SMILES 
cross-entropy, and torsion GCE losses over the selected positions.

\subsubsection{Downstream Fine-tuning}\label{downstream_finetuning_overview}

For downstream tasks, the molecular graph encoders and Transformer are fine-tuned end-to-end with task-specific prediction heads. 
In all settings, a learnable \texttt{[CLS]} token is prepended to the input sequence, and its final hidden representation is used for prediction.

\paragraph{Molecular Property Prediction}

For molecular property prediction, the input consists of motif tokens derived from the 2D molecular graph and SMILES tokens. 
No torsion tokens are used during fine-tuning.
The graph encoders and Transformer are initialized from the Stage II masked token modeling pretraining and fine-tuned jointly. 
The final hidden representation of the \texttt{[CLS]} token is passed to a task-specific linear head to predict molecular properties.

\paragraph{Drug-Target Interaction Prediction}

For DTI prediction, each protein sequence is encoded using the pretrained
ProtBERT model (\texttt{Rostlab/prot\_bert})~\cite{elnaggar2021prottrans}.
Protein sequences longer than 1,022 residues are truncated from the
C-terminal end so that the complete input, including two special tokens,
does not exceed the maximum length of 1,024 tokens. Final-layer residue
embeddings are extracted after excluding special and padding tokens and
mean-pooled to obtain a 1,024-dimensional protein representation. The
ProtBERT parameters remain frozen during embedding extraction.

The pooled protein representation is projected into the shared
300-dimensional space through a learnable linear layer and introduced as
a single protein token alongside the molecular motif tokens. Protein and
motif tokens are jointly processed by the Transformer, and the resulting
\texttt{[CLS]} representation is used for task-specific prediction.

\subsection{Training and Implementation Details}

Stage I is trained for 100 epochs using AdamW with a learning rate of 
$1\times10^{-3}$, a weight decay of $1\times10^{-2}$, and a batch size 
of 256. During Stage II, the graph encoders are frozen, and the remaining 
components are trained for 100 epochs using AdamW with a learning rate of 
$5\times10^{-5}$, a weight decay of $1\times10^{-2}$, and a batch size 
of 512. A linear learning-rate schedule with 10\% warmup is used in 
Stage II.

For DTI pretraining, the model is trained on BindingDB affinity data for 
100 epochs using AdamW without weight decay and a batch size of 512. Mean 
squared error is used as the training loss. The molecular encoders and 
protein projection layer use a learning rate of $1\times10^{-3}$, whereas 
the Transformer uses $1\times10^{-4}$.

All downstream MoTIF-X models are trained for 100 epochs with a batch
size of 512 using AdamW without weight decay.
Classification and regression models are optimized using binary
cross-entropy with logits and mean squared error, respectively.
Transformer dropout is set to 0 during Stage II pretraining and
regression fine-tuning, 0.2 for molecular classification, and 0.1
for DTI classification.
No early stopping is applied to downstream MoTIF-X training.

For OpenADMET regression, checkpoints are selected by the lowest
validation MAE.
For the supplementary MoleculeNet experiments, checkpoints are
selected by the highest validation ROC-AUC for classification
and the lowest validation mean squared error for regression.
DTI checkpoints are selected by the highest validation AUPR
for classification and Pearson correlation for regression.

All MoTIF-X models are trained on a single NVIDIA H100 graphics
processing unit.
For molecular property fine-tuning, the graph encoders use a
learning rate of $5\times10^{-4}$ and the Transformer uses
$1\times10^{-4}$.
For DTI classification, the molecular encoders and protein
projection layer use $1\times10^{-3}$, and the Transformer uses
$1\times10^{-4}$.

\subsection{Datasets}\label{method_datasets}

Additional details and statistics for the multimodal pretraining
and DTI regression datasets are provided in Supplementary Section B.

\paragraph{Multimodal Pretraining Data}

Multimodal pretraining is conducted using molecules with 3D conformations from the GEOM-Drugs dataset~\cite{axelrod2022geom}, which contains approximately 304,000 drug-like molecules with multiple conformers. Following prior work~\cite{stark20223d}, the five lowest-energy conformers are selected for each molecule during pretraining. Each selected conformer is treated as a separate training instance, with 
the corresponding motif and canonical SMILES tokens shared across 
conformers.

\paragraph{Molecular Property Prediction}

OpenADMET ExpansionRx~\cite{openadmet_expansionrx2026} is used as
the primary benchmark for molecular property prediction.
The benchmark comprises nine ADME regression endpoints:
LogD, KSOL, HLM CLint, MLM CLint, Caco-2 Papp, Caco-2 Efflux,
MPPB, MBPB, and MGMB.
Each endpoint is modeled separately using molecules with available
labels for that endpoint.

Additional evaluations are conducted on the MoleculeNet
benchmarks~\cite{wu2018moleculenet}, including BBBP, Tox21,
Toxicity Forecaster (ToxCast), Side Effect Resource (SIDER),
ClinTox, Maximum Unbiased Validation (MUV), HIV, the
beta-secretase 1 inhibitor activity (BACE) dataset, ESOL,
and Lipophilicity.
These experiments provide supplementary comparisons across
classification and regression tasks related to toxicity,
biological activity, and physicochemical properties.

\paragraph{DTI Regression}

For DTI regression, we construct an IC$_{50}$ dataset from BindingDB~\cite{liu2007bindingdb}. Dataset construction and preprocessing follow the Therapeutics Data Commons benchmark protocol~\cite{Huang2021tdc}, retaining high-confidence binding measurements for single-protein targets.

Before constructing the pretraining splits, all compound-protein pairs appearing in the downstream DTI classification benchmarks were removed from the BindingDB affinity dataset. Pair identity was determined using 
the combination of canonical SMILES and target protein sequence. This filtering ensured that the supervised DTI pretraining data were pair-disjoint from all downstream DTI classification benchmarks.

For external evaluation, we use ChEMBL version 35~\cite{zdrazil2024chembl} to construct a drug cold-start dataset of compound-protein pairs with experimentally measured IC$_{50}$ values, which are converted to pIC$_{50}$. Compounds present in the BindingDB training set are excluded, ensuring that all compounds in the external dataset are unseen during training.

\paragraph{DTI Classification}

For DTI classification, we follow the benchmark datasets and data preparation protocol introduced in MolTrans~\cite{huang2021moltrans}. 
Experiments are conducted on the BIOSNAP, BindingDB, and DAVIS datasets, together with the Unseen Drugs and Unseen Targets generalization settings.
Dataset construction, train/validation/test splits, and negative sampling procedures strictly follow the protocol described in~\cite{huang2021moltrans}.

\subsection{Evaluation Protocol}\label{evaluation_protocol}

For OpenADMET ExpansionRx, the official training and test partitions
are used without transferring molecules between them.
For each endpoint, 10\% of the labeled official training partition
is randomly sampled as the validation set using a fixed splitting
seed of 42, and the remaining samples are used for training.
The resulting train, validation, and test assignments are fixed
across all compared methods and downstream random seeds.
The official test partition is not used for model selection.

Label scales and performance metrics follow the official OpenADMET
ExpansionRx evaluation protocol~\cite{openadmet_expansionrx2026}.
LogD is modeled and evaluated on its original scale.
For the remaining endpoints, labels are transformed as
$\log_{10}(\max(y,0)+1)$ before training.
Predictions for these endpoints are clipped to a minimum of zero
on the transformed scale during validation and testing.
Training predictions are not clipped.
Endpoint performance is evaluated using mean absolute error (MAE).
Relative absolute error (RAE) is calculated as the MAE divided
by the mean absolute deviation of the test labels from their mean.
Overall performance is summarized by macro-averaged relative
absolute error (MA-RAE), the equally weighted average of RAE
across the nine endpoints.

For the supplementary MoleculeNet evaluations, a deterministic,
chirality-aware Bemis-Murcko scaffold split~\cite{bemis1996properties}
is used with an 80:10:10 train/validation/test ratio.
MoTIF-X and all baseline models use identical training, validation,
and test partitions, which remain fixed across all random seeds.
Classification and regression performance are evaluated using
ROC-AUC and root mean squared error (RMSE), respectively.

For BindingDB affinity regression, compound-protein pairs are randomly
partitioned at the pair level into training, validation, and test sets
using a 70:10:20 ratio and a fixed random seed of 42.
This split does not enforce disjoint compounds or proteins across subsets.
The ChEMBL drug-cold-start dataset is used exclusively for external
evaluation and contains no compounds present in the BindingDB training set.
DTI classification datasets use their predefined benchmark splits.
DTI regression is evaluated using PCC.
DTI classification is evaluated using AUPR, computed as average
precision, following the MolTrans protocol~\cite{huang2021moltrans}.

Model selection is based exclusively on validation performance.
The selected checkpoint is evaluated once on the corresponding test set
after training.
Molecular and DTI pretraining are each performed once, and the resulting
checkpoints are reused across five downstream runs with different
random seeds.
Endpoint results are reported as the mean and standard deviation
of individual-model scores across these five runs.

\paragraph{Statistical Comparisons}

For statistical comparisons, all methods are trained using the same
set of five random seeds.
Predictions from these five downstream runs are averaged for each
test sample to form an ensemble prediction for each method.
Test samples are aligned by molecular identity for property prediction
and by drug-protein pair identity for DTI classification,
with matching reference labels.
Statistical comparisons assess differences between five-seed ensembles.
Performance summaries report the mean and standard deviation of
individual-model scores across the same five seeds.

For OpenADMET, two-sided paired sign-flip permutation tests are
performed on per-molecule absolute-error differences between
MoTIF-X and each comparator using 10,000 permutations.
Holm correction is applied separately to the 81 baseline comparisons
across nine endpoints and nine baselines, and the 72 ablation
comparisons across nine endpoints and eight configurations.

For DTI classification, differences in ensemble AUPR between
MoTIF-X and each baseline are assessed using two-sided paired
stratified bootstrap tests with 10,000 replicates.
Positive and negative test samples are resampled separately,
and identical resampling indices are applied to both methods.
Holm correction is applied jointly to the 25 comparisons across
five benchmark settings and five baselines.

Statistical significance is defined as a Holm-adjusted
$p$-value below 0.05.
Detailed comparison procedures and results are provided
in Supplementary Section A.

\bibliography{main}% common bib file

\section*{Abbreviations}

\begin{longtable}{@{}p{0.20\textwidth}p{0.74\textwidth}@{}}
\toprule
\textbf{Abbreviation} & \textbf{Definition} \\
\midrule
\endfirsthead

\toprule
\textbf{Abbreviation} & \textbf{Definition} \\
\midrule
\endhead

\bottomrule
\endfoot

2D & two-dimensional \\
3D & three-dimensional \\
ADMET & absorption, distribution, metabolism, excretion, and toxicity \\
AUPR & area under the precision-recall curve \\
BACE & beta-secretase 1 inhibitor activity dataset \\
BBBP & blood-brain barrier penetration \\
BIOSNAP & Stanford Biomedical Network Dataset Collection \\
\texttt{[CLS]} & classification token \\
ClinTox & clinical toxicity dataset \\
DTI & drug-target interaction \\
ECFP & extended-connectivity fingerprint \\
ESOL & estimated solubility dataset \\
GCE & Gaussian cross-entropy \\
GIN & Graph Isomorphism Network \\
GNN & graph neural network \\
HCC & hepatocellular carcinoma \\
HIV & human immunodeficiency virus \\
IC$_{50}$ & half-maximal inhibitory concentration \\
IKBKB & inhibitor of nuclear factor kappa-B kinase subunit beta \\
MAE & mean absolute error \\
MA-RAE & macro-averaged relative absolute error \\
MLP & multilayer perceptron \\
MUV & Maximum Unbiased Validation \\
PCC & Pearson correlation coefficient \\
pIC$_{50}$ & negative base-10 logarithm of the molar IC$_{50}$ value \\
RAE & relative absolute error \\
RMSE & root mean squared error \\
ROC-AUC & area under the receiver operating characteristic curve \\
SIDER & Side Effect Resource \\
SMILES & simplified molecular-input line-entry system \\
STAT3 & signal transducer and activator of transcription 3 \\
Tox21 & Toxicology in the 21st Century \\
ToxCast & Toxicity Forecaster \\
UMAP & uniform manifold approximation and projection \\
\end{longtable}

\section*{Declarations}

\subsection*{Availability of data and materials}

The source code for MoTIF-X, together with the molecular property
and DTI classification benchmark data used in this study, is
available at \url{https://github.com/Bin-Chen-Lab/Motif-X}.
The OpenADMET ExpansionRx dataset~\cite{openadmet_expansionrx2026}
is available at
\url{https://huggingface.co/datasets/openadmet/openadmet-expansionrx-challenge-data}.
The GEOM-Drugs molecular pretraining data are available from
Harvard Dataverse at
\url{https://dataverse.harvard.edu/dataset.xhtml?persistentId=doi:10.7910/DVN/JNGTDF}.
The BindingDB IC$_{50}$ data were obtained through Therapeutics
Data Commons at
\url{https://tdcommons.ai/multi_pred_tasks/dti/}.
ChEMBL version 35 is available from the ChEMBL release archive at
\url{https://ftp.ebi.ac.uk/pub/databases/chembl/ChEMBLdb/releases/chembl_35/}.
The STAT3 structure used for molecular docking is available from
the Protein Data Bank under accession code 6QHD.

\subsection*{Competing interests}

The authors declare that they have no competing interests.

\subsection*{Funding}

This work was supported by the National Institutes of Health 
(R01GM145700 and R61HL177451), the National Institute on Aging of the 
National Institutes of Health (R01AG072449), the National Science Foundation 
(IIS-2212174), and the Michigan State University Strategic Partnership Grant.

\subsection*{Authors' contributions}

L.M., J.Z. and B.C. conceived the study. L.M. developed the MoTIF-X 
framework, implemented the models, constructed the datasets, performed the 
experiments, analyzed the results, prepared the figures, and wrote the 
initial manuscript. J.Z. and B.C. jointly supervised the study, provided 
conceptual guidance, contributed to the interpretation of the results, 
reviewed the implementation, and revised the manuscript. All authors read 
and approved the final manuscript.

\subsection*{Acknowledgements}

Not applicable.

\section*{Additional files}

\noindent
\textbf{File name:} Additional file 1.pdf\\
\textbf{File format:} PDF (.pdf)\\
\textbf{Title:} Supplementary Information for ``MoTIF-X: A Multimodal 
Tokenized Framework for Interpretable and Extensible Molecular 
Representation Learning''\\
\textbf{Description:} Supplementary benchmark results, dataset statistics, 
baseline descriptions, attention and representation analyses, motif-level 
attribution analyses, library-scale STAT3 screening, and molecular docking 
methods and results.

\end{document}